\documentclass{article}
\usepackage{spconf,amsmath,graphicx}
\usepackage[pagebackref,breaklinks,colorlinks,citecolor=cvprblue]{hyperref}

\usepackage{amssymb,amsfonts}%AMS extra symbols and fonts
\usepackage{dblfloatfix}%fix double column figure ordering and placement

\usepackage[ruled,vlined]{algorithm2e}
\usepackage{graphicx}

\usepackage{booktabs}
\usepackage{siunitx}
\usepackage[numbers,compress]{natbib}
\usepackage{texnames}
\usepackage{bm,bbm}
\usepackage{orcidlink}
\usepackage{multirow}
\usepackage{xcolor}
\definecolor{background_color}{RGB}{255,255,255}  % Alice Blue

\definecolor{cnn_color}{RGB}{230,230,255}  % Alice Blue
\definecolor{transformer_color}{RGB}{255,230,240}  % Floral White
\definecolor{ours_color}{RGB}{230,255,235}  % Mint Cream
\definecolor{input_color}{RGB}{255,240,240}      % 淺灰色
\definecolor{mamba_color}{RGB}{235,235,225}  % Floral White

\usepackage{colortbl}  % 用於表格顏色
\definecolor{cvprblue}{rgb}{0.21,0.49,0.74}
\usepackage[pagebackref,breaklinks,colorlinks,citecolor=cvprblue]{hyperref}

\newcommand{\best}[1]{\textcolor{bestColor}{\textbf{#1}}}      % 最佳
\newcommand{\secondBest}[1]{\textcolor{secondBestColor}{\textbf{#1}}} % 次佳
\definecolor{orange}{RGB}{255,107,0}

\definecolor{bestColor}{RGB}{255, 0, 0}    % 最佳 - 紅色
\definecolor{secondBestColor}{RGB}{0, 0, 255} % 次佳 - 藍色
\definecolor{thirdBestColor}{RGB}{0, 100, 0}  % 第三佳 - 深綠色
\definecolor{darkgreen}{rgb}{0.0, 0.5, 0.0} % 定義深綠色 (R,G,B)

\begin{document}
\title{HSSDCT: Factorized Spatial-Spectral Correlation for Hyperspectral Image Fusion
}
%\name{Anonymous Authors}
\name{Chia-Ming Lee$^{1,2}$ \quad Yu-Hao Ho$^3$ \quad Yu-Fan Lin$^1$ \quad Jen-Wei Lee$^1$ \quad Li-Wei Kang$^3$ \quad Chih-Chung Hsu$^{1,2}$}

\address{$^1$Institute of Data Science, National Cheng Kung University\\$^2$Institute of Intelligent Systems, National Yang Ming Chiao Tung University\\$^3$Department of Electrical Engineering, National Taiwan Normal University}

\maketitle
\begin{abstract}
Hyperspectral image (HSI) fusion aims to reconstruct a high-resolution HSI (HR-HSI) by combining the rich spectral information of a low-resolution HSI (LR-HSI) with the fine spatial details of a high-resolution multispectral image (HR-MSI). Although recent deep learning methods have achieved notable progress, they still suffer from limited receptive fields, redundant spectral bands, and the quadratic complexity of self-attention, which restrict both efficiency and robustness. To overcome these challenges, we propose the Hierarchical Spatial-Spectral Dense Correlation Network (HSSDCT). The framework introduces two key modules: (i) a Hierarchical Dense-Residue Transformer Block (HDRTB) that progressively enlarges windows and employs dense-residue connections for multi-scale feature aggregation, and (ii) a Spatial-Spectral Correlation Layer (SSCL) that explicitly factorizes spatial and spectral dependencies, reducing self-attention to linear complexity while mitigating spectral redundancy. Extensive experiments on benchmark datasets demonstrate that HSSDCT delivers superior reconstruction quality with significantly lower computational costs, achieving new state-of-the-art performance in HSI fusion. Our code is available at  \hyperlink{https://github.com/jemmyleee/HSSDCT}{https://github.com/jemmyleee/HSSDCT}.
\end{abstract}


\section{Introduction}\label{sec1}

Hyperspectral imaging (HSI) provides abundant spectral information across contiguous bands, enabling fine-grained material identification and supporting applications in areas such as environmental monitoring, precision agriculture, and medical diagnostics. However, simultaneously acquiring high spatial and high spectral resolution (HR-HSI) is difficult due to hardware limitations and high acquisition costs, especially on platforms such as satellites or UAVs. As a result, HSI fusion (also called pansharpening) has emerged as a practical solution, reconstructing HR-HSI by combining the rich spectral information of a low-resolution HSI (LR-HSI) with the spatial details of a high-resolution multispectral image (HR-MSI). 

%Driven by deep learning, HSI fusion has achieved notable progress beyond traditional model-driven approaches \cite{17,18}. 
Early networks employed 3D-CNNs for spectral-spatial HSI fusion \cite{20}, residual architectures such as PZRes-Net \cite{xu2021progressive}, or UNet-based variants like Dual-UNet \cite{jiang2021dual}. Other works explored blind HSI fusion \cite{27}, optimization priors \cite{zhang2022deep}, or model interpretability \cite{29}. More recently, Transformer-based designs \cite{lee2025robust_igarss,lee2025hyfusion_igarss} such as FusFormer \cite{fusformer} and HyperTransformer \cite{HyperTransformer} have been introduced to capture long-range dependencies, while hybrid frameworks including HyperRefiner \cite{HyperRefiner}, QRCODE \cite{QRCODE}, U2Net \cite{U2Net}, and FusionMamba \cite{FusionMamba} further enhance spatial-spectral modeling.  

Despite these advances, several challenges remain. Many recent methods attempt to enlarge the receptive field for capturing long-range spatial-spectral dependencies, for example, by introducing multi-scale convolutions and large kernel operations, or attention/state-space modules. While these strategies improve the ability to model global correlations, they often suffer from quadratic computational complexity, making the use of large windows or long sequences impractical in real applications. Moreover, most existing designs treat spatial and spectral information jointly in a single attention formulation, which not only increases redundancy but also fails to exploit the unique structure of hyperspectral data explicitly. As a result, models are computationally heavy, less efficient, and more vulnerable to noise from redundant bands. Balancing receptive field enlargement, efficiency, and redundancy mitigation, therefore, remains a core challenge in HSI fusion.  

To address these issues, we propose the Hierarchical Spatial-Spectral Dense Correlation Network (HSSDCT), an efficient architecture tailored for high-fidelity HSI fusion. At its core, HSSDCT employs a Hierarchical Dense-Residue Transformer Block (HDRTB), which progressively enlarges windows across layers and integrates dense-residue connections to enable the enlarged-receptive-field features while promoting stable feature propagation. Afterwards, we introduce the Spatial-Spectral Correlation Layer (SSCL), which explicitly factorizes aggregation into spatial and spectral sub-pathways. This design both mitigates spectral redundancy by modeling band-wise relationships separately, reducing the quadratic cost of self-attention to linear complexity, enabling effective long-range modeling at a lower computational burden. Through judicious integration of HDRTB and SSCL, HSSDCT achieves superior fusion
quality with improved efficiency. Experiments confirm that the proposed model attains state-of-the-art performance while remaining lightweight, making it more suitable for real-world deployment.

\section{PROPOSED METHOD}
\label{sec:method}

Our goal is to reconstruct an HR-HSI by leveraging the complementary information of an LR-HSI and an HR-MSI. To achieve this, we design HSSDCT, a lightweight yet practical architecture built upon two core ideas: (i) hierarchical feature extraction with dense-residue connections, and (ii) efficient spatial-spectral correlation modeling.

\subsection{Overall Framework}
\begin{figure}[t]
    \centering
    \includegraphics[width=\linewidth]{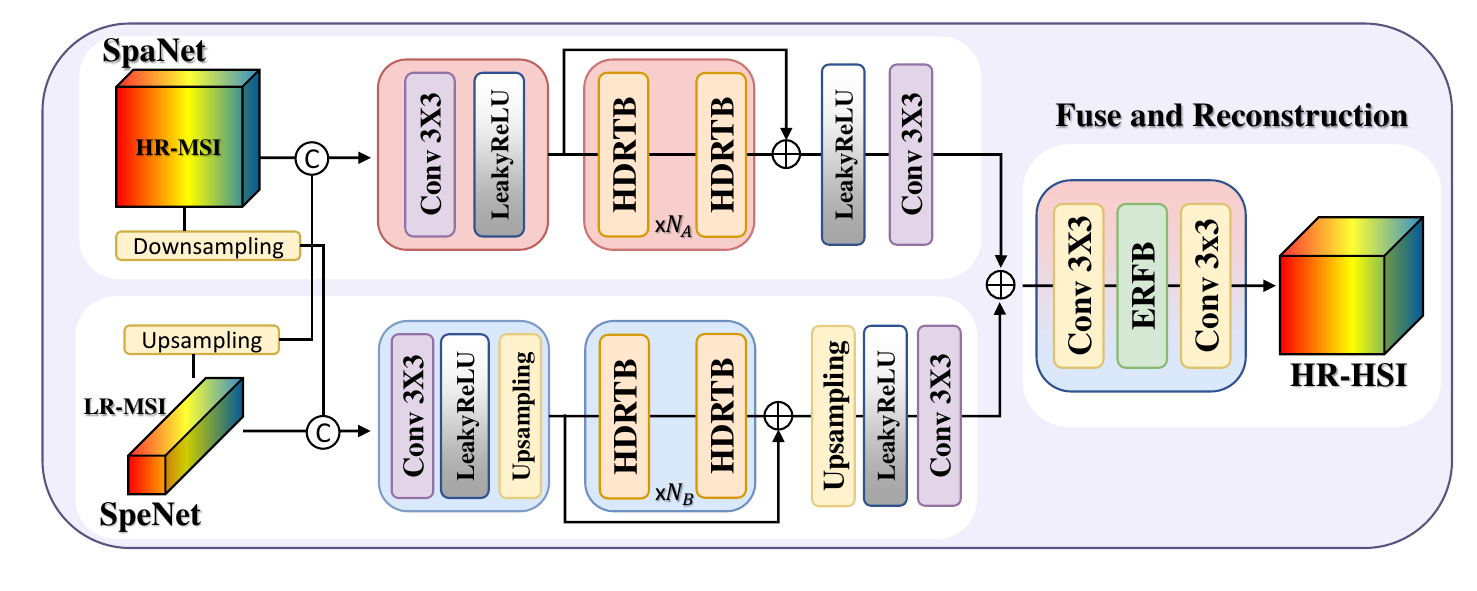}
    \caption{Overview of the proposed HSSDCT framework. The network takes an LR-HSI and an HR-MSI as inputs. The outputs of two parallel branches are finally fused to reconstruct the HR-HSI.}
    \label{fig:flowchart} 
\end{figure}

As illustrated in Figure \ref{fig:flowchart}, HSSDCT takes an LR-HSI $\mathbf{X}_{h}$ and an HR-MSI $\mathbf{X}_{m}$ as inputs. The LR-HSI, containing rich spectral information but coarse spatial details, is processed by the spectral branch, while the HR-MSI, with fine spatial structures but limited bands, is processed by the spatial branch. Both branches first apply shallow convolutions, followed by stacked HDRTB modules. Their outputs are fused through element-wise addition and refinement layers to generate the reconstructed HR-HSI $\mathbf{Y}^{*}$ effectively. This dual-branch design ensures that complementary spectral and spatial features are jointly exploited for high-quality fusion. Formally, the final output can be expressed as:
\begin{equation}
\mathbf{Y}^{*} = \mathcal{F}_{\text{final}}\big(\mathbf{F}_{\text{Spe}} + \mathbf{F}_{\text{Spa}}\big),
\end{equation}
where $\mathbf{F}_{\text{Spe}}$ and $\mathbf{F}_{\text{Spa}}$ denote the features extracted by the spectral and spatial branches, and $\mathcal{F}_{\text{final}}(\cdot)$ represents the feature fusion and image reconstruction layers for desired HR-HSI.

\begin{figure*}[t]
    \centering
    \includegraphics[width=\textwidth]{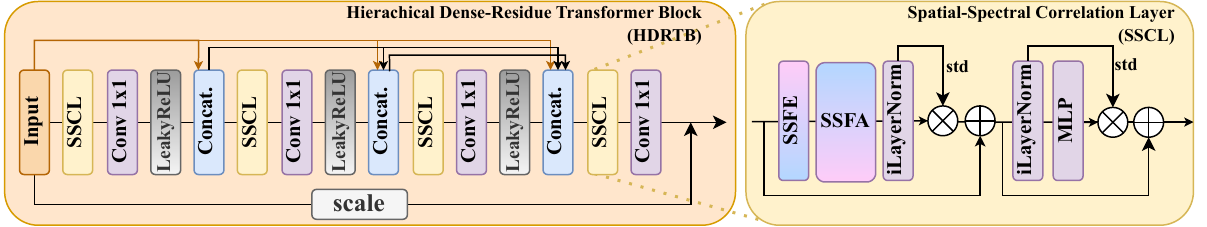}
    \caption{HSSDCT integrates the HDRTB for enlarged-receptive-field feature extraction using progressively enlarged windows, and the SSCL for efficient factorized spatial-spectral correlation modeling with linear complexity.}
    \label{fig:detail}
\end{figure*}
\vspace{-0.8mm}

\subsection{Hierarchical Dense-Residue Transformer Block}
The HDRTB is designed to serve as the backbone for spatial-spectral feature extraction in HSSDCT, as shown in Figure \ref{fig:detail}. 

First, \textbf{hierarchical windows} are employed within the block: rather than using a fixed attention window size, successive layers operate on progressively enlarged windows (e.g., $h_i \times w_i$ increasing with depth) \cite{zhang2024hitsr}. This design allows the network to capture fine local textures in early layers while gradually expanding the receptive field to aggregate broader spatial-spectral context in deeper layers \cite{10204559}, which is crucial for recovering global and semantic structures of the reconstructed HR-HSI. 

\begin{figure}[t!]
\centering
\includegraphics[width=1\linewidth]{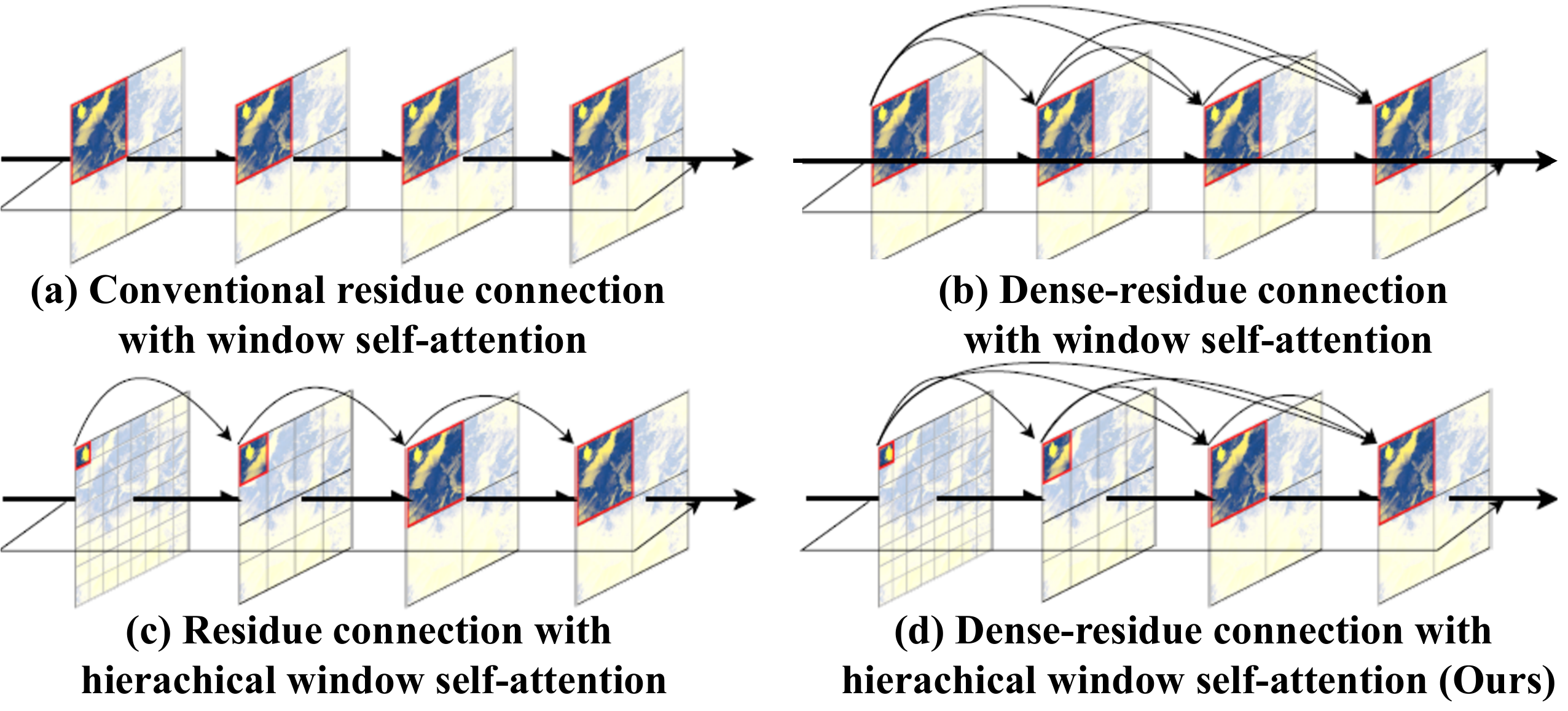}\vspace*{-0.3cm}
\caption{\small Hierarchical Dense-Residue Transformer Block (HDRTB) with hierarchical windows and dense connections, effectively enlarging the receptive field of features.}
\label{fig:efficient}
\end{figure}

Second, \textbf{dense-residue connections} are integrated to greatly improve the effective receptive fields without increasing complexity significantly \cite{Hsu_2024_CVPR}. Features from all layers are concatenated and projected through $1\times1$ convolutions before being passed to the next layer. A final residual skip connection further stabilizes training and preserves original information. This ensures that each HDRTB layer benefits from both shallow and deep representations, avoiding vanishing gradients and enhancing feature reuse, as shown in Figure \ref{fig:efficient}. Formally, the block output can be expressed as:
\begin{equation}
\mathbf{F}_{out} = \mathbf{F}_{in} + \gamma \cdot \text{Conv}_{1\times1}\big(\text{Cat}(\mathbf{F}_{1}, \mathbf{F}_{2}, \mathbf{F}_{3})\big),
\end{equation}
where $\gamma$ is a scaling factor for structure preserving, which is empirically set to $0.2$. By jointly leveraging hierarchical windows and dense-residue connections, HDRTB achieves efficient multi-scale feature aggregation with an enlarged effective receptive field compared to conventional Transformer blocks.

\begin{figure}
    \centering
    \includegraphics[width=0.5\textwidth]{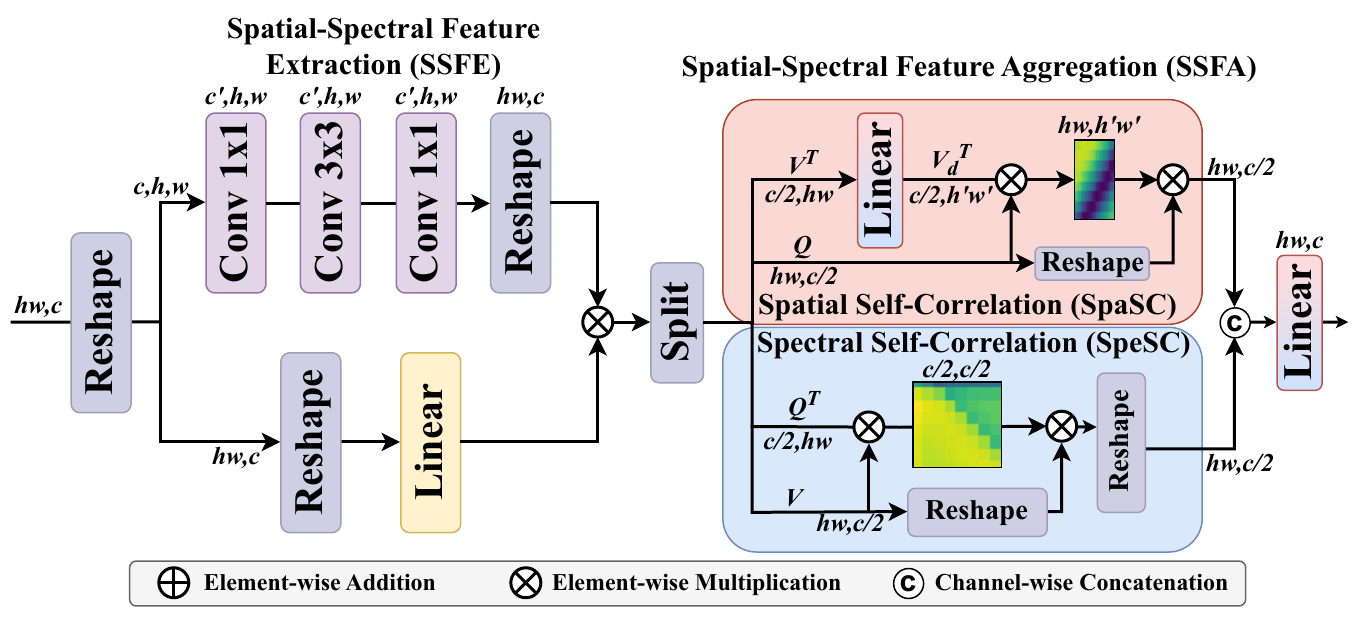}
    \vspace{-0.75cm}   
    \caption{Spatial-Spectral Correlation Layer (SSCL) factorizes spatial and spectral correlations for efficient aggregation, reducing quadratic attention cost while preserving both spatial context and spectral fidelity.}

    \label{fig:SSC} 
\end{figure}

\subsection{Spatial-Spectral Correlation Layer}
Computational complexity for HSI tasks is also essential. The proposed SSCL decouples feature aggregation into spatial and spectral correlation pathways with \textbf{linear complexity}, thereby enabling the use of large hierarchical windows inside HDRTB. Besides, adaptive layer normalization \cite{Peebles2022DiT}, noted with iLayerNorm, is embedded in SSCLs.

\textbf{Feature Projection.} Given an input feature map $\mathbf{F}\in\mathbb{R}^{C\times H\times W}$, the Spatial-Spectral Feature Extraction (SSFE) module generates query and value representations. Unlike common self-attention that applies identical linear projections, SSFE employs two lightweight branches: a convolutional path for local spatial encoding and a linear path for spectral encoding by channel splitting for matrix factorization. Their fusion produces $\mathbf{Q}$ and $\mathbf{V}$ that are enriched with both spatial and spectral contexts with reduced complexity. Details are shown in Figure \ref{fig:detail} and \ref{fig:SSC}. 

\textbf{Spatial Self-Correlation (SpaSC).} Within each hierarchical window, SpaSC computes spatial affinities by correlating queries with spatially compressed value tokens, enabling efficient context modeling across varying window sizes:
\begin{equation}
\text{SpaSC}(\mathbf{Q},\mathbf{V}) = \frac{\mathbf{Q}\mathbf{V}^T}{\sqrt{d}} \mathbf{V}.
\end{equation}
This formulation captures long-range spatial dependencies without the prohibitive cost of quadratic self-attention.

\textbf{Spectral Self-Correlation (SpeSC).} Complementary to SpaSC, SpeSC models correlations across the spectral dimension. By computing channel-wise affinities, it preserves fine-grained spectral signatures critical for hyperspectral fidelity:
\begin{equation}
\text{SpeSC}(\mathbf{Q},\mathbf{V}) = \frac{\mathbf{Q}^T\mathbf{V}}{HW}\mathbf{V}^T.
\end{equation}

\textbf{Feature Aggregation.} The outputs of SpaSC and SpeSC are fused through element-wise addition and followed by the Spatial-Spectral Feature Aggregation (SSFA). This yields spatial-spectral representations that combine broad contextual awareness with spectral integrity.

%By factorizing spatial and spectral correlations, SSCL removes the quadratic bottleneck of standard attention, allowing HDRTB to employ progressively larger windows. This design significantly enlarges the effective receptive field while maintaining computational efficiency, making HSSDCT scalable and practical.

%\textbf{Advantages.} 
Compared to conventional window self-attention, SSCL offers three key benefits: (i) reduced complexity that scales linearly with window size, (ii) ability to employ progressively larger windows in HDRTB for broader receptive field features, and (iii) explicit modeling of spectral correlations, which are often overlooked in Transformer-based HSI fusion methods. Together, HDRTB and SSCL form the core of HSSDCT, striking a balance between accuracy and efficiency for high-quality HSI fusion.

\section{Experimental Results}
\label{sec:expALL}

\subsection{Experiment Settings}

\begin{table*}[ht]
\centering
\caption{{\textbf{Performance evaluation and complexity comparison of the proposed HSSDCT and other HSI fusion methods.}} The best results are highlighted in \best{bold-red}, while the second-best results are highlighted in \secondBest{bold-blue}. Rows are colored to distinguish different approaches: \colorbox{cnn_color}{CNN-based}, \colorbox{transformer_color}{Transformer-based}, \colorbox{mamba_color}{Mamba-based}, and \colorbox{ours_color}{the proposed} methods.} 
\scalebox{0.65}{
\begin{tabular}{r|cccc|cccc|cccc}  
\toprule[0.15em]
\multirow{2}{*}{\textbf{Method}} &  \multicolumn{4}{c|}{Model Complexity} & \multicolumn{4}{c|}{4 Bands HR-MSI} & \multicolumn{4}{c}{6 Bands HR-MSI}\\
 &  \#Params↓ & FLOPs↓ & Run-time↓ & Memory↓ & PSNR↑ &  SAM↓ &  RMSE↓  & ERGAS↓  &  PSNR↑ &  SAM↓ &  RMSE↓ & ERGAS↓   \\\hline
\rowcolor{cnn_color}
\textbf{PZRes-Net} \cite{xu2021progressive} {\tt\small{TIP'21}} & 40.15M & 5262.34G & 0.0141s & 11059MB & 34.963 &  1.934  & 35.498 & 1.935& 37.427 &  1.478 &  28.234 &  1.538  \\
\rowcolor{cnn_color}
\textbf{MSSJFL} \cite{Min2021MSSJFL} {\tt\small{HPCC'21}} & 16.33M & 175.56G & 0.0128s & \secondBest{1349MB} &  34.966 & 1.792 &  33.636 &  2.245 &  38.006 &  1.390 &  26.893 & 1.535 \\
\rowcolor{cnn_color}
\textbf{Dual-UNet} \cite{Xiao2021DualUNet} {\tt\small{TGRS'21}} & \secondBest{2.97M} & \secondBest{88.65G} & \secondBest{0.0127s} & 2152MB & \secondBest{35.423} & 1.892 & 33.183 & \secondBest{1.796} & 38.453 & 1.548 & 26.148 & 1.205 \\
\rowcolor{cnn_color}
\textbf{DHIF-Net} \cite{zhang2022deep} {\tt\small{TCI'22}} & 57.04M & 13795.11G &  6.005s & 14936MB &34.458 & 1.829 & 34.769 & 2.613 & \secondBest{39.146} & 1.239 & 25.309 & \secondBest{1.113} \\
\rowcolor{transformer_color}
\textbf{FusFormer} \cite{fusformer} {\tt\small{TGRS'22}} & \best{0.18M} & \best{11.74G} & 0.0158s & 5964MB & 34.217 & 2.012 & 35.687 & 1.996 & 38.637 & 1.678 & 28.674 & 1.204 \\
\rowcolor{transformer_color}
\textbf{HyperTransformer} \cite{HyperTransformer} {\tt\small{CVPR'22}} & 142.83M & 343.96G & 0.0252s & 8104MB & 28.692 &  3.664 &  62.231 & 4.774 & 32.954 & 2.568 & 41.256 & 3.834 \\
\rowcolor{transformer_color}
\textbf{HyperRefiner} \cite{HyperRefiner} {\tt\small{TJDE'23}} & 19.32M & 94.37G & 0.0237s & 7542MB & 33.298 & 2.129 & 38.769 & 2.086 & 37.654 & 1.590 & 29.629 & 1.403  \\
\rowcolor{cnn_color}
\textbf{U2Net} \cite{U2Net} {\tt\small{ACMMM'23}} & 265.15M & 1931.09G & 0.1684s & 7506MB & 25.622 & 3.855 & 86.682 & 7.341 & 27.068 & 3.832 & 85.101 & 6.816 \\
\rowcolor{cnn_color}
\textbf{PSDNet} \cite{gong2023multipatch} {\tt\small{TGRS'23}} & 3.155M & 663.004G & 0.0354s & 3962MB & 35.153 & 1.967 & 64.573 & 1.834 & 38.588 & 1.619 & 52.446 & 1.088 \\
\rowcolor{cnn_color}
\textbf{QRCODE} \cite{QRCODE} {\tt\small{TGRS'24}} & 41.88M & 2231.19G & 0.2452s & 15028MB & 35.361 & \secondBest{1.623} & \secondBest{32.711} & 2.027 & 38.948 & \secondBest{1.148} &  \secondBest{24.617} & 1.429 \\ 
\rowcolor{mamba_color}
\textbf{FusionMamba} \cite{FusionMamba} {\tt\small{TGRS'24}} & 21.68M & 134.47G & 0.0347s & 2446MB & 30.741 & 1.978 &  50.744 & 2.945 & 32.407 & 1.540 & 45.774 & 2.569 \\
%\rowcolor{transformer_color}
%\textbf{HyFusion} \cite{HyFusion} {\tt\small{IGARSS'25}} & 25.53M & 641.49G &  0.0294s & 2572MB & \best{37.421} &  \best{1.256} &  \best{27.425} &  \best{1.172}& \best{40.197} & \best{1.017} &  \best{21.455} &  \best{1.076}  \\
\hline
\rowcolor{ours_color}
\textbf{HSSDCT (Ours)}& 6.78M & 283.84G &  \best{0.0112s} & \best{1290MB} & \best{37.212} &  \best{1.348} &  \best{26.544} &  \best{1.648}& \best{40.227} & \best{1.098} &  \best{20.524} &  \best{1.059}  \\
\bottomrule[0.15em]
\end{tabular}}
\label{tab:performance}
\end{table*}
\begin{figure*}[htbp]
\centering
\includegraphics[width=0.92\linewidth]{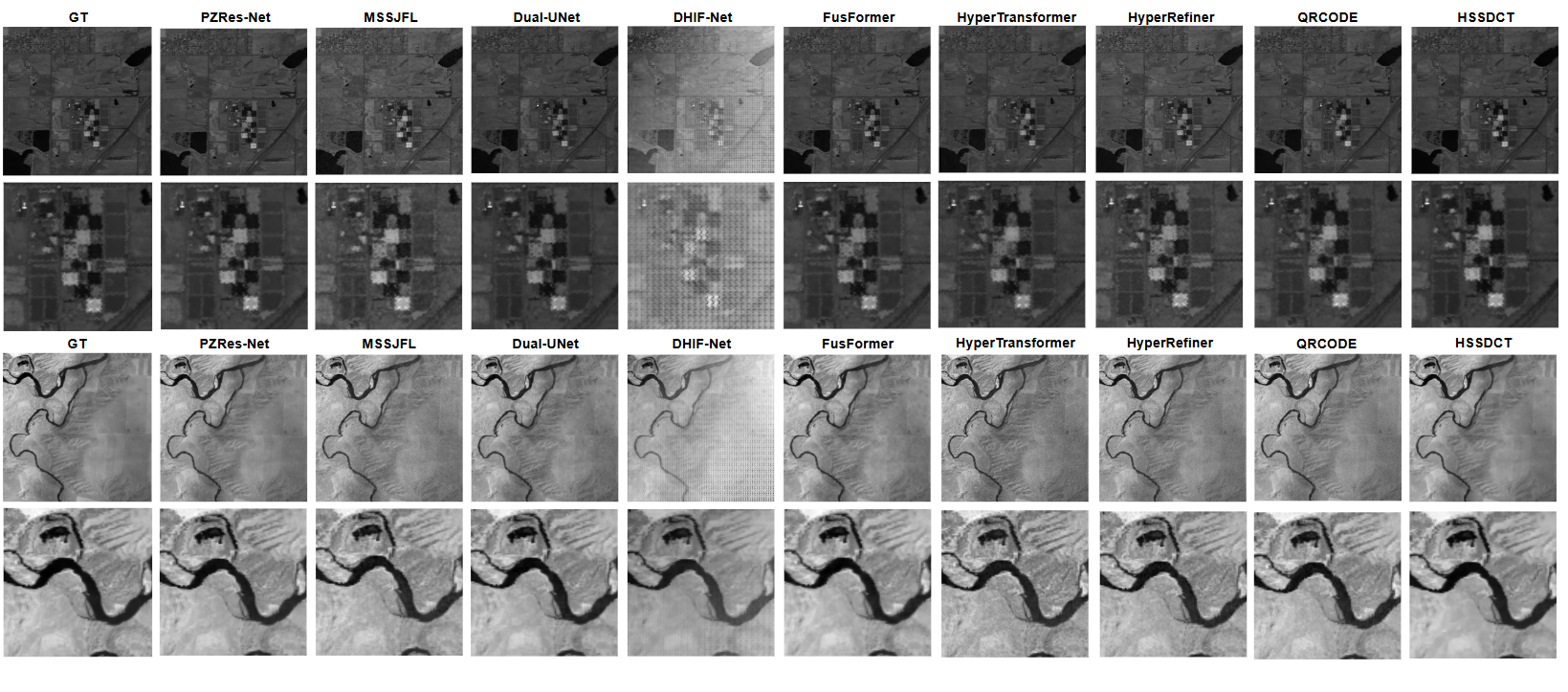}
\vspace{-5mm}
\caption{\small Visualization of fused results and residue images for LR-HSI and HR-MSI inputs, comparing HSSDCT with representative baselines.}
\label{fig:exp1}
\vspace{-5mm}
\end{figure*}
\textbf{Dataset Illustration.}
The dataset used for performance evaluation in this study was acquired by the Airborne Visible/Infrared Imaging Spectrometer (AVIRIS) sensor \cite{Vane1993airborne}. It consisted of 2,078 HR-HSI images, which were randomly partitioned into training, validation, and testing sets for performance evaluation. The training set contained 1,678 images, while the validation and testing sets contained 200 images for each. The spatial and spectral resolutions of the HR-MSI and LR-HSI were $256\times 256 \times M_m$ and $64\times 64\times 172$, respectively, where $M_m$ is either 4 or 6 in our experiments.

\textbf{Implementation Details.} The proposed method was implemented using the PyTorch deep learning framework and NVIDIA-GeForce RTX 3090. The batch size was set to 4, and the number of training epochs was fixed to 600 for all experiments involving the proposed method. For the peer methods, the number of training epochs was set according to their default values as specified in their respective original publications. The ADAM \cite{Kingma2014adam} optimizer was used for training, with an initial learning rate of 0.0001. The learning rate was adjusted during the training process using the Cosine Annealing learning decay scheduler. The window size of HDRTBs is set to \{4,8,16,16\}.The network is trained with a composite loss combining pixel fidelity, spectral preservation, and frequency consistency:
\begin{equation}
\ell_{\text{Total}} = \ell_{\text{L1}} + \lambda_{1}\ell_{\text{SAM}} + \lambda_{2}\ell_{\text{SWT}},
\end{equation}
where $\ell_{\text{L1}}$ enforces pixel fidelity, $\ell_{\text{SAM}}$ (Spectral Angle Mapper) preserves spectral signatures by minimizing angular differences between spectra, and $\ell_{\text{SWT}}$ (Stationary Wavelet Transform) maintains structural textures via wavelet-domain consistency. We empirically set $\lambda_{1}, \lambda_{2}=0.01$.

\subsection{Quantitative Results}

To evaluate the performance of our method, we compared it with five other supervised HSI fusion methods: PZRes-Net \cite{Zhu2021PZRes}, MSSJFL \cite{Min2021MSSJFL}, Dual-UNet \cite{Xiao2021DualUNet}, DHIF-Net \cite{Huang2022DHIF}, FusFormer \cite{fusformer}, HyperTransformer \cite{HyperTransformer}, HyperRefiner \cite{HyperRefiner}, QRCODE \cite{QRCODE}, and FusionMamba \cite{FusionMamba}. The performance was objectively measured using PSNR, SAM, RMSE, and ERGAS. Experiments were conducted with both 4 and 6 HR-MSI bands. Table \ref{tab:performance} presents the quantitative comparison with state-of-the-art methods.

The results demonstrate that the proposed framework achieves better fusion quality compared to existing methods. HSSDCT consistently delivers the best results under both 4-band and 6-band settings, while maintaining the lowest runtime and memory usage, highlighting its strong balance between accuracy and efficiency. Moreover, the visual comparison in Fig.~\ref{fig:exp1} shows that our method preserves sharper spatial structures and more faithful spectral signatures, producing reconstructions closer to the ground truth.

\section{Conclusion}
\label{sec:conclusion}

We presented HSSDCT, a lightweight yet effective framework for HSI fusion. The proposed architecture addresses three major challenges: limited receptive fields, the quadratic cost of attention, and spectral redundancy. By introducing the HDRTB with hierarchical windows and dense-residue connections, and the SSCL with factorized spatial and spectral modeling, HSSDCT achieves efficient long-range feature extraction with reduced complexity. Extensive experiments confirm that HSSDCT delivers state-of-the-art fusion accuracy while maintaining a compact design, highlighting both its effectiveness and efficiency. These results demonstrate the potential of HSSDCT as a practical and scalable solution for real-world remote sensing or HSI applications.

\small
\bibliographystyle{IEEEtranN}
\bibliography{sn-bibliography}

\end{document}